\documentclass[letterpaper]{article} 
\usepackage{aaai2026}  
\usepackage{times}  
\usepackage{helvet}  
\usepackage{courier}  
\usepackage[hyphens]{url}  
\usepackage{graphicx} 
\usepackage{natbib}  
\usepackage{caption} 
\usepackage{amsmath}
\usepackage{algorithm}
\usepackage{algorithmic}
\usepackage{enumitem}
\usepackage{booktabs}
\usepackage{multirow}
\usepackage{graphicx}
\usepackage{subcaption}
\usepackage{booktabs}
\usepackage[table,xcdraw]{xcolor}
\usepackage{hhline} 
\usepackage{makecell} 
\usepackage{float}
\usepackage{tcolorbox}

\usepackage{amssymb}
\usepackage{amsfonts}
\usepackage{amsthm}
\usepackage{mathtools}
\usepackage{bbding}
\usepackage{multicol}
\usepackage{multirow}
\usepackage{scalerel}
\usepackage{amssymb}
\usepackage{subcaption}
\usepackage{color}
\usepackage{xcolor}
\usepackage{colortbl}
\usepackage{diagbox}
\usepackage{xpatch}

\usepackage{tablefootnote}

\makeatletter
\newif\ifdiagbox@cellEmpty@

\define@key{diagbox}{widest}{%
  \ifdiagbox@cellEmpty@
    \setkeys{diagbox}{innerwidth=\widthof{#1}}%
    \PackageWarning{diagbox}{innerwidth is set to \the\diagbox@wd}%
  \fi}

\define@key{diagbox}{highest}{%
  \ifdiagbox@cellEmpty@
    \setkeys{diagbox}{height=#1}%
  \fi}

\define@key{diagbox}{text}{%
  \def\diagbox@text{#1}}

\define@key{diagbox}{align}{%
  \in@{#1}{l, c, r}%
  \ifin@
    \def\diagbox@align{#1}%
  \else
    \PackageError{diagbox}%
      {The value of key "text" must be one of "l", "c", and "r".}%
  \fi}

\xpatchcmd{\diagbox@double}{%
  \setkeys{diagbox}{dir=NW,#1}%
}{%
  \if\relax\detokenize{#2}\relax
    \if\relax\detokenize{#3}\relax
      \diagbox@cellEmpty@true
      \setkeys{diagbox}{highest=1\line, align=l, text=\@empty}%
    \fi
  \fi
  \setkeys{diagbox}{dir=NW, #1}%
  \ifdiagbox@cellEmpty@
    \rlap{\makebox
      [\dimexpr\diagbox@wd-\diagbox@insepl-\diagbox@insepr\relax]%
      [\diagbox@align]%
      {\diagbox@text}}%
  \fi
}{}{\ddt}
\makeatother

\newfloat{figtab}{htb}{fgtb}
\makeatletter
  \newcommand\figcaption{\def\@captype{figure}\caption}
  \newcommand\tabcaption{\def\@captype{table}\caption}
\makeatother

\newtheorem{rmk}{Remark}

\usepackage{pifont}

\usepackage{amsmath,amsfonts,bm}
\usepackage{amsthm}

\def\eqref#1{equation~(\ref{#1})}

\def\1{\bm{1}}

\DeclareMathAlphabet{\mathsfit}{\encodingdefault}{\sfdefault}{m}{sl}
\SetMathAlphabet{\mathsfit}{bold}{\encodingdefault}{\sfdefault}{bx}{n}

\newcommand{\R}{\mathbb{R}}

\usepackage{newfloat}
\usepackage{listings}
\DeclareCaptionStyle{ruled}{labelfont=normalfont,labelsep=colon,strut=off} 
\floatstyle{ruled}
\newfloat{listing}{tb}{lst}{}
\floatname{listing}{Listing}
\title{FedGRPO: Privately Optimizing Foundation Models with Group-Relative Rewards from Domain Clients}
\author{
    Gongxi Zhu\textsuperscript{\rm 1},
    Hanlin Gu\textsuperscript{\rm 2},
    Lixin Fan\textsuperscript{\rm 2},
    Qiang Yang\textsuperscript{\rm 3},
    Yuxing Han\textsuperscript{\rm 1}\thanks{Corresponding author.}
    \\
}
\affiliations{
    \textsuperscript{\rm 1}Shenzhen International Graduate School, Tsinghua University\\
    \textsuperscript{\rm 2}AI Lab, Webank\\
    \textsuperscript{\rm 3} Department of Data Science and Artificial Intelligence, The Hong Kong Polytechnic University

    gx.zhu@foxmail.com, allengu@webank.com, yuxinghan@sz.tsinghua.edu.cn
}

\usepackage{bibentry}

\begin{document}
\maketitle


\begin{abstract}

One important direction of Federated Foundation Models (FedFMs) is leveraging data from small client models to enhance the performance of a large server‑side foundation model. Existing methods based on model level or representation level knowledge transfer either require expensive local training or incur high communication costs and introduce unavoidable privacy risks. We reformulate this problem as a reinforcement learning style evaluation process and propose FedGRPO, a privacy preserving framework comprising two modules. The first module performs competence-based expert selection by building a lightweight confidence graph from auxiliary data to identify the most suitable clients for each question. The second module leverages the “Group Relative” concept from the Group Relative Policy Optimization (GRPO) framework by packaging each question together with its solution rationale into candidate policies, dispatching these policies to a selected subset of expert clients, and aggregating solely the resulting scalar reward signals via a federated group–relative loss function. By exchanging reward values instead of data or model updates, FedGRPO reduces privacy risk and communication overhead while enabling parallel evaluation across heterogeneous devices. Empirical results on diverse domain tasks demonstrate that FedGRPO achieves superior downstream accuracy and communication efficiency compared to conventional FedFMs baselines.

%

\end{abstract}
\begin{links}
    \link{Code}{https://github.com/Liar-Mask/FedGRPO}
\end{links}
\section{Introduction}


Federated Foundation Models (FedFMs) \cite{fan2025ten,kang2023grounding,ren2025advances} present a promising paradigm that integrates the strong generalization capabilities of server-side Foundation Models (FMs) with the domain-specific expertise of client devices.  One important goal in FedFMs is how to effectively leverage clients' domain knowledge to enhance the performance of FMs while preserving the privacy of local data. 

Current methods for integrating domain knowledge from downstream clients into a server-side foundation model (FM) fall into two main categories: model-level transfer \cite{fan2023fate, zhang2023fedpetuning} and synthetic data-level transfer \cite{yu2023multimodal,abacha2024synthetic}. In model-level transfer, each client fine-tunes a server-provided small portion of model parameters or an adapter on its local data and sends the updated parameters back for aggregation. In synthetic data-level transfer, clients generate domain-pertinent synthetic data, which is then uploaded to the server to enhance the FM’s performance. Both approaches impose substantial communication overhead and expose sensitive information: the frequent exchange of parameters or synthetic data not only strains network bandwidth \cite{zhao2024disentangling} but also invites semi-honest adversaries to infer private data from the transmitted artifacts \cite{chen2024unveiling}.

In order to reduce privacy leaking risk and communication overhead, we take a novel approach by which only {model evaluation scores} are transferred from clients to the server.  Compared with model parameters or synthetic data, evaluation scores are orders of magnitude smaller in terms of amount of information to be transmitted, and moreover, the risk of privacy leaking is substantially reduced.
In this framework, the server leverages evaluations from domain-specific clients on distributed problems and candidate solutions to simultaneously enhance model performance while preserving data privacy. \textit{A fundamental requirement of this framework is that client evaluations should be accurate.} Specifically, it introduces two critical challenges: (1) 
how to select the most suitable clients for evaluating specific problems, based on their domain expertise;
and (2) how to aggregate evaluations from multiple expert clients to effectively improve the foundation model.

To address these challenges, we propose FedGRPO, a federated-foundation-models (FedFMs) framework that integrates two complementary modules: (i) competence-based expert selection and (ii) Group‑Relative Reward Aggregation.
The first module constructs a confidence graph from auxiliary data to quantify each client’s expertise on domain-specific queries, enabling the server to recruit the most competent subset of experts for every request.
Building on "Group Relative" idea of Group Relative Policy Optimization (GRPO), the second module computes relative rewards among the selected experts to guide server-side updates. Concretely, the server maintains a global policy and periodically dispatches enhanced candidate policies—each encapsulating the original question alongside a detailed problem-solving rationale—to the chosen expert clients. Each client evaluates the received policy on its private in-domain data and returns only a scalar reward signal, thereby preserving data privacy. The server then aggregates these signals through a federated group-relative loss function, which balances contributions from multiple experts and iteratively refines the foundation model.
FedGRPO offers three key advantages: (1) it leverages domain-specific user data to improve LLM training; (2) it enhances data privacy by exchanging only reward signals rather than raw data or model parameters; and (3) it improves computational efficiency by enabling parallel reward evaluation across multiple Clients. Our contributions are summarized as follows:
\begin{itemize}
    \item We introduce FedGRPO, a novel reinforcement-learning-inspired FedFM pipeline that recasts large-model refinement as a reward-based evaluation process. By packaging each query with its solution rationale into candidate policies and leveraging a lightweight confidence graph for competence-based expert selection, FedGRPO enables efficient, parallelized evaluation across resource‑constrained clients without requiring expensive local fine-tuning.
    \item We design a federated Group Relative Policy Optimization module that aggregates only scalar reward signals from selected expert clients via a group-relative loss. This mechanism eliminates the need to transmit raw data, model updates, or high-dimensional representations—substantially reducing communication overhead and mitigating privacy leakage.
    \item Extensive experiments show that FedGRPO: 1) enhances server-model reasoning using client reward feedback and group-relative loss, closely matching centralized GRPO performance; 2) effectively leverages heterogeneous domain expertise through competence-based expert selection even without ground-truth answers; and 3) avoids privacy leakage risks inherent in model-level or synthetic data-level transfer while maintaining smaller orders of magnitude  communication overhead.
\end{itemize}

\section{Related Work}
\label{sec: related work}

\subsection{Knowledge Transfer in FedFMs}

Federated Foundation Models (FedFMs) \cite{fan2025ten,kang2023grounding,ren2025advances}  constitute a distributed learning paradigm facilitating the reciprocal exchange and adaptation of knowledge between server-hosted Foundation Models (FMs) and the domain-specific expertise residing on client devices.
Existing methodologies for client-to-server knowledge transfer within the FedFMs framework can be broadly bifurcated into two primary categories: model-level and data-level transfer.

In the context of model-level transfer, the substantial scale of foundation models precludes clients from training the entire model locally for subsequent uploading and aggregation. To circumvent this limitation, researchers have extended existing Parameter-Efficient Fine-Tuning (PEFT) techniques to the federated learning setting, a methodology termed FedPEFT \cite{sun2024improving, yi2023pfedlora,zhang2023fedpetuning}. Within this framework, clients perform lightweight PEFT on the FM locally and subsequently transmit only the trained PEFT modules or adapters to the server. These components are then aggregated and integrated with the global FM.

Regarding data-level transfer schemes, a prevalent approach involves clients generating domain-pertinent synthetic data \cite{li2024federated,abacha2024synthetic,hou2025private}. This synthetic data encapsulates domain knowledge while ostensibly preserving the privacy of the original dataset from which it was derived.The synthesized data is then uploaded to the server to enhance the FM's performance through further training. Such schemes can be augmented with differential privacy mechanisms to fortify privacy preservation.

Notwithstanding the efficacy of the aforementioned schemes in transferring domain-specific knowledge from clients to the server, the frequent transmission of adapters or synthetic data imposes considerable communication overhead \cite{zhao2024disentangling} and presents inherent risks of privacy leakage \cite{chen2024unveiling}.

\subsection{Reinforcement Learning for LLM Reasoning}


Advances in LLM research have shifted focus from basic autoregressive token generation to complex reasoning tasks like mathematical problem-solving and code generation. Reinforcement learning (RL) enhances model reasoning through trial-and-error optimization, emerging as a key method for post-training LLMs \cite{chu2025sft}.
Proximal Policy Optimization (PPO) \cite{schulman2017proximal} is a well-known traditional RL method that is widely used due to its robustness across various fields. However, it encounters challenges such as the need for costly online data collection, high computational overhead, and sensitivity to hyperparameters. Subsequently, offline RL methods, such as Direct Preference Optimization (DPO) \cite{rafailov2023direct}, were proposed to eliminate the need for live interaction or data generation and allow users to tune the model more efficiently. Deepseek developed Group Relative Policy Optimization (GRPO) \cite{shao2024deepseekmath} method, which abandons the value model in favor of estimating the baseline based on group scores, significantly reducing training resource requirements. The advantages of this approach were further emphasized with its application on Deepseek-R1.

\section{The Proposed Method}

\begin{figure*}[!t]
    \centering
    \includegraphics[scale=0.8]{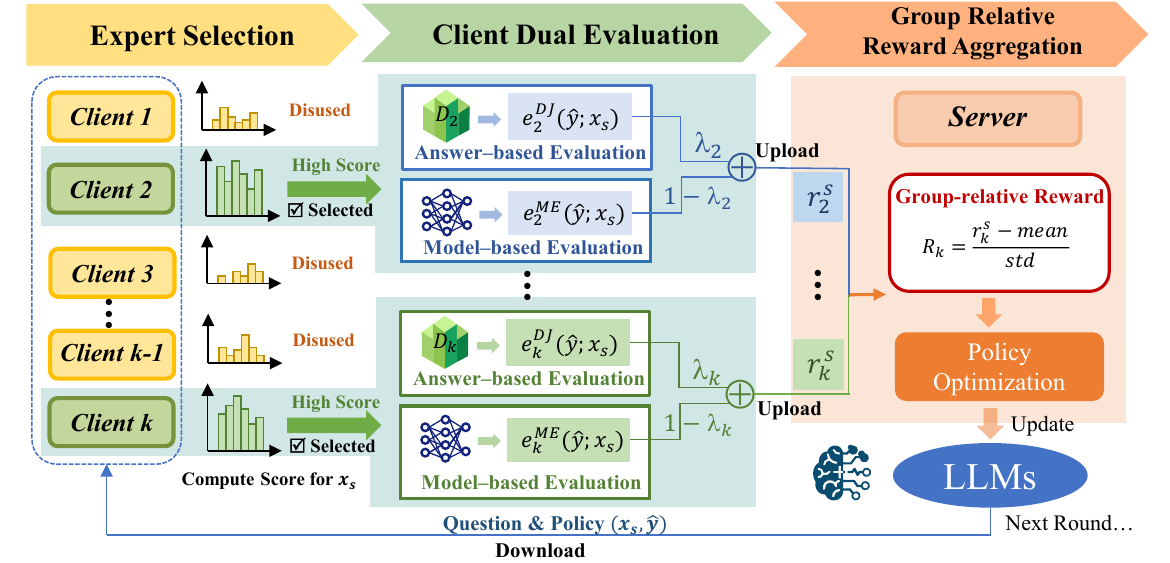}
    \kern-10pt
    \caption{Overview of FedGRPO including three steps: 1) Expert selection to select an appropriate expert subset $\mathcal{C}(x_{s})$ for every question $x_{s}$; 2) Dual evaluation on the select client $ k \in \mathcal{C}(x_{s})$ to compute rewards $r_k^s$ for the policy $\hat{y}$; and 3) Group relative reward aggregation on server to get group-relative reward $R_{k}$ and perform policy optimization to update LLMs.}
    \label{Fig:framwork}
    \kern-10pt
\end{figure*}
\subsection{Problem Formulation}
\paragraph{Setting.}
We consider a collaborative learning framework comprising a central server and $K$ distributed clients. Each client $k \in \{1, \dots, K\}$ possesses a local dataset $\mathcal{D}_k = \{(x_i^{(k)}, y_i^{(k)})\}_{i=1}^{n_k}$, where $x_i^{(k)} \in \mathcal{X}$ represents the input instance and $y_i^{(k)} \in \mathcal{Y}$ denotes the corresponding output. For instance, $x_i^{(k)}$ can be a question and $y_i^{(k)}$ its solution.  

Each client trains a model on its domain-specific dataset $\mathcal{D}_k$ to have their domain-specific model $\theta_k$. The central server hosts a large-scale pre-trained foundation model, denoted by $\theta_g$, with the goal of leveraging domain knowledge from clients’ private datasets. Suppose the central server holds only a negligible amount of auxiliary data, denoted by 
$\displaystyle\mathcal{D}_{p}
              =\{(x_{p,j},\,y_{p,j})\}_{j=1}^{|\mathcal{D}_{p}|}$. The server aims to optimize:
\begin{equation}
    \min_{\theta_g} \ell(\theta_g, \mathcal{D}_1, \dots, \mathcal{D}_K, \mathcal D_p),
\end{equation}
to enhance the utility of $\theta_g$ on domain-specific data. However, privacy constraints prohibit the server from directly accessing clients’ raw data. 

\paragraph{Threat model.} We assume an honest-but-curious threat model, where the central server and participating clients follow the prescribed protocol but may attempt to infer additional information from the received messages. Existing approaches typically transmit models, gradients, or embeddings rather than raw data to the server; nevertheless, these strategies remain susceptible to privacy leakage \cite{zhu2019deep}.

\paragraph{Formulation.} Motivated by reinforcement learning (RL), we propose to treat each client’s model as a reward model that evaluates the server’s generated outputs. In this design, clients provide only scalar reward signals to the server, thereby preventing the leakage of raw data.  
We formulate the server’s optimization as an RL problem. Specifically, the server maintains a policy $\pi_{\theta_g}$, parameterized by $\theta_g$, which, given an input $x \in \mathcal{X}$ (e.g., a question), generates a candidate output $\hat{y} \in \mathcal{Y}$ (e.g., an answer or solution):
\begin{equation}
    \hat{y} \sim \pi_{\theta_g}(\cdot | x).
\end{equation}

Each client $k$ evaluates the generated output $\hat{y}$ with respect to its local domain knowledge or ground-truth labels in $\mathcal{D}_k$, and returns a scalar reward:
\begin{equation}
    r_k(\hat{y}, x) = f_{\theta_k}(\hat{y}; \mathcal{D}_k),
\end{equation}

where $f_{\theta_k}$ denotes the reward function that derives reward score from the local data or client’s locally trained model $\theta_k^*$ on dataset $\mathcal{D}_k$.  The server aggregates individual client rewards $\{r_k(\hat{y}, x)\}_{k=1}^K$ through an aggregation function $A(\cdot)$, resulting in a global reward $R(\hat{y} | x)$. Consequently, the server's optimization objective can be expressed as:
\begin{equation}
\begin{split}
        J(\theta_g) &= \mathbb{E}_{x \sim \mathcal{X}, \, \hat{y} \sim \pi_{\theta_g}(\cdot | x)} \Big[ R(\hat{y} | x) \Big]. \\
                &= \mathbb{E}_{x \sim \mathcal{X}, \, \hat{y} \sim \pi_{\theta_g}(\cdot | x)} \Big[ A(\{r_k(\hat{y}, x)\}_{k=1}^K) \Big].
\end{split}
\end{equation}
However, this formulation poses two key challenges: (1) determining how to select suitable clients for evaluating specific tasks, considering the heterogeneity in their domain expertise; and (2) developing effective strategies to aggregate evaluations from multiple expert clients to enhance the performance of the foundation model.



\subsection{FedGRPO}

To tackle the aforementioned dual challenges, we propose \textsc{FedGRPO}, which integrates two tightly coupled modules: \emph{expert client selection} and \emph{group-relative policy optimization}~(GRPO). The expert‐selection module estimates each client’s competency on the auxiliary data and selects an appropriate subset of expert clients for every question; the GRPO module then transforms the raw scores provided by experts into a unified, scale-invariant reinforcement signal.

\paragraph{Competence-based expert selecting.}
For each unlabeled question (no answer) $x_{s}\!\sim\!\mathcal{D}_{s}$, the server  
(i) samples a provisional answer
$\hat{y}\!\sim\!\pi_{\theta_g}(\!\cdot\!\mid x_{s})$ and  
(ii) embeds the question with a frozen encoder
$\phi$ derived from the foundation model, producing
$\mathbf z_{s}=\phi(x_{s})\in \R^{d}$.  
Based on the cosine similarity between $\mathbf z_{s}$ and auxiliary embeddings
$\{\mathbf z_{p,j}\}_{j=1}^{|\mathcal{D}_{p}|} = \{\phi(x_{p,j})\}_{j=1}^{|\mathcal{D}_{p}|}$, the server retrieves the
$L$ most similar labeled exemplars
\begin{equation}
  \label{eq:nbhd}
  \begin{split}
      \mathcal{G}(x_{s})=\bigl\{(x_{p,\ell},y_{p,\ell})\bigr\}_{\ell=1}^{L}, \\
  \qquad
  \langle\mathbf z_{s}, \mathbf z_{p,1}\rangle\ge\dots\ge
  \langle\mathbf z_{s},\mathbf z_{p,L}\rangle .
    \end{split}
\end{equation}
The server distributes $\mathcal G(x_s)$ to all clients evaluate. Each client maintains a running accuracy estimate on auxiliary prompts; the
server therefore possesses a \emph{competence score}
$r_{k}^{p}(x_{s})\in[0,1]$ for every client~$k$ with respect to the
neighborhood~\eqref{eq:nbhd} as:
\[
   r_{k}^{p}
   =\frac{1}{L}\sum_{\ell=1}^{L}
     \mathbf 1\!\bigl[
        e_{k}(y_{p,\ell};x_{p,\ell}) = y_{p,\ell}
     \bigr].
\]  
It retains the
$M$ highest-scoring experts for each question $x_s$:
\begin{equation}
  \label{eq:experts}
  \mathcal{C}(x_{s})=\operatorname*{Top\mbox{-}M}_{k}\bigl\{r_{k}^{p}(x_{s})\bigr\}.
\end{equation}

\begin{algorithm}[H]
\caption{\textsc{FedGRPO} training with competence-based expert selection}
\label{alg:FedGRPO}
\textbf{Input:} auxiliary prompt set $\mathcal D_p$, unlabeled queries $\mathcal D_s$, global policy $\pi_{\theta_g}$, client set $\mathcal K$; neighbourhood size $L$, experts per question $M$; learning rate $\eta$, epochs $T$
\begin{algorithmic}[1]
\STATE Pre-compute $\ell_2$-normalised embeddings $\mathbf z_{p,j}=\phi(x_{p,j})$ for all $(x_{p,j},y_{p,j})\!\in\!\mathcal D_p$ 
\FOR{$t=1$ \TO $T$}                     
    \STATE Sample question $x_s \sim \mathcal D_s$ and provisional answer $\hat y \sim \pi_{\theta_g}(\cdot\mid x_s)$
    \STATE Compute question embedding $\mathbf z_s=\phi(x_s)$
    \STATE Retrieve $L$ nearest auxiliary exemplars $\mathcal G(x_s)=\{(x_{p,\ell},y_{p,\ell})\}_{\ell=1}^L$ 
    \STATE \textbf{Broadcast} $\mathcal G(x_s)$ to all clients $k\!\in\!\mathcal K$
    \FOR{clients $k\!\in\!\mathcal K$ \textbf{in parallel}}
        \STATE Compute competence score $r_k^{p}(x_s)=\frac{1}{L}\sum_{\ell=1}^{L}\mathbf 1\!\bigl[e_k(y_{p,\ell};x_{p,\ell})=y_{p,\ell}\bigr]$
        \STATE \textbf{Send} $r_k^{p}(x_s)$ to server
    \ENDFOR
    \STATE Select experts $\mathcal C(x_s)=\operatorname*{Top\text{-}M}_k\{r_k^{p}(x_s)\}$   \COMMENT{Eq.~\eqref{eq:experts}}
    \STATE \textbf{Broadcast} $\langle x_s,\hat y\rangle$ to experts $k\!\in\!\mathcal C(x_s)$
    \FORALL{experts $k\!\in\!\mathcal C(x_s)$ \textbf{in parallel}}
        \STATE Choose pathway $\lambda_k(x_s)\!\in\!\{0,1\}$ (AE if ground-truth exists, else ME)
        \STATE Compute private score $r_k^{s}= \lambda_k e_k^{\mathrm{AE}}(\hat y;x_s) + (1-\lambda_k) e_k^{\mathrm{ME}}(\hat y;x_s)$
        \STATE \textbf{Send} $r_k^{s}$ to server
    \ENDFOR
    \STATE Compute group-relative reward 
    \begin{equation}
        \begin{split}
                  & \mu_r=\frac{1}{M}\sum_{k\in\mathcal C}r_k^{s},\quad
           \sigma_r=\sqrt{\tfrac{1}{M}\sum_{k\in\mathcal C}(r_k^{s}-\mu_r)^2},  \\ 
           & R_{k}=\frac{r_k^{s}-\mu_r}{\sigma_r+\epsilon}    
        \end{split}
    \end{equation}
    \STATE Update global policy
           $\theta_g \gets \theta_g + \eta\,R_{k}\,
           \nabla_{\theta_g}\log\pi_{\theta_g}(\hat y\mid x_s)$
\ENDFOR
\end{algorithmic}
\end{algorithm}

\paragraph{Dual evaluation on the clients.}
For each training iteration,
the server broadcasts the $\langle x_{s},\hat{y}\rangle$ solely to the selected expert set $\mathcal{C}(x_{s})$.
Upon receiving the triplet
$\langle x_{s},\hat{y},\mathcal{G}(x_{s})\rangle$,
each client~$k$ produces two scalar feedback signals:
\emph{private score} $r_{k}^{s}$ for the unlabeled question
and \emph{auxiliary score}~$r_{k}^{p}$ for its retrieved neighbourhood.
The precise mechanism depends on which evaluation criteria the
client can apply.

\begin{enumerate}[label=(\roman*), leftmargin=2.2em]
\item \textbf{Answer–based evaluation (AE).}
      If the question $x_{s}$ (or a paraphrase thereof) appears in the
      client’s private corpus $\mathcal{D}_{k}$ together with a trusted
      ground-truth answer $\tilde{y}_{k}(x_{s})$, the client performs an
      \emph{exact-answer check}:
      \[
          e_{k}^{\mathrm{AE}}(\hat{y};x_{s}) \;=\;
          \mathbf 1\!\bigl[\,\hat{y}= \tilde{y}_{k}(x_{s})\,\bigr].
      \]
      The resulting score is binary and therefore directly comparable
      across all AE clients.

\item \textbf{Model–based evaluation (ME).}
      If the question is \emph{not} covered by $\mathcal{D}_{k}$, the client
      falls back to its self-trained reward model
      $f_{\theta_k^\star}$, yielding a real-valued score
      \[
          e_{k}^{\mathrm{ME}}(\hat{y};x_{s})
          \;=\;
          f_{\theta_k^\star}(\hat{y};x_{s})\in\R.
      \]
      Typical choices for $f_{\theta_k^\star}$ include a small
      cross-entropy classifier or a learned rubric rubricator.
\end{enumerate}

\noindent
In practice a single client may support \emph{both}
criteria and dynamically select the appropriate pathway via a gating
indicator
$\lambda_{k}(x_{s})\!\in\!\{0,1\}$:
\begin{equation*}
       e_{k}(\hat{y};x_{s})
   \;=\;
   \lambda_{k}(x_{s})\,e_{k}^{\mathrm{AE}}(\hat{y};x_{s})
   \;+\;
   \bigl(1-\lambda_{k}(x_{s})\bigr)\,
   e_{k}^{\mathrm{ME}}(\hat{y};x_{s}).
\end{equation*}

The private score is then $r_{k}^{s}=e_{k}(\hat{y};x_{s})$.
This dual-evaluation design lets every client exploit the strongest
knowledge source at its disposal (ground-truth answers when available,
otherwise a learned evaluator) while providing the server with a
question-specific competence signal $r_{k}^{p}$ and a raw reward
$r_{k}^{s}$ suitable for GRPO aggregation.
\begin{rmk}
We also incorporate a format reward to ensure that the generated answers conform to the expected type, following the approach in \cite{shao2024deepseekmath}.
\end{rmk}

\paragraph{Group Relative Policy Optimization.}

The server retains only the $M$ highest-scoring clients
\[
  \mathcal{C}(x_{s})=\operatorname*{Top\text{-}M}_{k}\{\,r_{k}^{p}\},
\]
disregarding the rest.  
This question-adaptive pruning ensures that subsequent optimization relies on the most competent—and therefore most reliable—experts.
Let $\{r_{k}^{s}\}_{k\in\mathcal{C}}$ be the private scores returned by the selected experts.  
Define
\begin{gather}
   \mu_r=\frac{1}{M}\sum_{k\in\mathcal{C}}r_{k}^{s},\! \quad \sigma_r=\sqrt{\frac{1}{M}\sum_{k\in\mathcal{C}}(r_{k}^{s}-\mu_r)^{2}}, \\
   R_k(x_{s},\hat{y})=\frac{r_{k}^{s}-\mu_r}{\sigma_r+\epsilon}.
\end{gather}



Standardising in this manner eliminates scale discrepancies across evaluation modes and dampens outliers, yielding the group-relative reward $R_{k}$.

Finally, the server performs a policy-gradient step
\[
  \theta_g \;\leftarrow\;
  \theta_g + \eta\,
  R_{k}(x_{s},\hat{y})\,
  \nabla_{\theta_g}\log\pi_{\theta_g}(\hat{y}\mid x_{s}),
\]
thus reinforcing responses that outperform expert-group average and steadily improving $\pi_{\theta_g}$ on unlabeled server data.

\begin{rmk}
In this work, the term “group” denotes the aggregation of reward signals from multiple clients and corresponds to the ensemble of candidate policies in GRPO \cite{shao2024deepseekmath}. 
To facilitate understanding, we use $\hat{y}$ to denote a single policy, whereas it could represents a collection of policies. 
\end{rmk}

\section{Experimental Results}
\label{sec: results}


\begin{table*}[!ht]
\centering
\footnotesize
\renewcommand{\arraystretch}{0.95}
\begin{tabular}{cclccccccc}
\toprule 
\rowcolor[HTML]{E2EFDA} 
\cellcolor[HTML]{D0CECE}Dataset & \cellcolor[HTML]{D0CECE}Model & \multicolumn{1}{c}{\cellcolor[HTML]{D0CECE}Method} & Math & Minerva & AMC & Olympiad & AIME24 & AIME25 & \cellcolor[HTML]{FFF2CC}Avg. \\ 
\midrule

 &  & Zero-shot & 0.316 & 0.081 & 0.272 & 0.203 & 0.074 & 0.034 & 0.163 \\
 &  & Fedpetuning+GRPO & 0.342 & 0.121 & 0.279 & 0.233 & 0.072 & 0.048 & 0.183 \\
 &  & Fedpetuning+SFT & 0.480 & 0.088 & 0.294 & 0.220 & 0.068 & 0.038 & 0.198 \\
 &  & DPSDA-FL+GRPO & 0.630 & 0.275 & 0.285 & 0.067 & \textbf{0.310} & \textbf{0.083} & 0.275 \\
 &  & DPSDA-FL+SFT & 0.498 & 0.129 & 0.284 & 0.260 & 0.079 & 0.049 & 0.217 \\
 &  & Central-GRPO & 0.701 & 0.307 & 0.440 & \textbf{0.353} & 0.081 & 0.056 & 0.323 \\
 & \multirow{-7}{*}{\begin{tabular}[c]{@{}c@{}}Qwen2.5-\\ Math-1.5B\end{tabular}} & \cellcolor[HTML]{DDEBF7}FedGRPO & \cellcolor[HTML]{DDEBF7}\textbf{0.716} & \cellcolor[HTML]{DDEBF7}\textbf{0.313} & \cellcolor[HTML]{DDEBF7}\textbf{0.451} & \cellcolor[HTML]{DDEBF7}0.348 & \cellcolor[HTML]{DDEBF7}0.133 & \cellcolor[HTML]{DDEBF7}0.065 & \cellcolor[HTML]{DDEBF7}\textbf{0.338} \\
\cmidrule(lr{0.5em}){2-10}

 &  & Zero-shot & 0.118 & 0.092 & 0.063 & 0.046 & 0.006 & 0.006 & 0.055 \\
 &  & Fedpetuning+GRPO & 0.384 & 0.129 & 0.156 & 0.132 & 0.017 & 0.006 & 0.137 \\
 &  & Fedpetuning+SFT & 0.318 & 0.085 & 0.100 & 0.061 & 0.006 & 0.003 & 0.096 \\
 &  & DPSDA-FL+GRPO & 0.448 & 0.148 & 0.099 & 0.176 & \textbf{0.029} & \textbf{0.009} & 0.152 \\
 &  & DPSDA-FL+SFT & 0.428 & 0.134 & 0.073 & \textbf{0.197} & 0.027 & 0.007 & 0.144 \\
 &  & Central-GRPO & \textbf{0.474} & \textbf{0.210} & 0.157 & 0.153 & 0.010 & 0.003 & \textbf{0.168} \\
 & \multirow{-7}{*}{Qwen2.5-3B} & \cellcolor[HTML]{DDEBF7}FedGRPO & \cellcolor[HTML]{DDEBF7}0.438 & \cellcolor[HTML]{DDEBF7}0.143 & \cellcolor[HTML]{DDEBF7}\textbf{0.195} & \cellcolor[HTML]{DDEBF7}0.157 & \cellcolor[HTML]{DDEBF7}0.024 & \cellcolor[HTML]{DDEBF7}0.005 & \cellcolor[HTML]{DDEBF7}0.160 \\
\cmidrule(lr{0.5em}){2-10}

 &  & Zero-shot & 0.426 & 0.121 & 0.326 & 0.163 & 0.111 & 0.048 & 0.199 \\
 &  & Fedpetuning+GRPO & 0.460 & 0.107 & 0.329 & 0.132 & 0.049 & 0.038 & 0.186 \\
 &  & Fedpetuning+SFT & 0.400 & 0.107 & 0.182 & 0.132 & 0.049 & 0.018 & 0.148 \\
 &  & DPSDA-FL+GRPO & 0.714 & \textbf{0.323} & 0.432 & 0.308 & 0.087 & 0.064 & 0.321 \\
 &  & DPSDA-FL+SFT & 0.574 & \textbf{0.323} & 0.343 & 0.187 & 0.067 & 0.026 & 0.253 \\
 &  & Central-GRPO & \textbf{0.742} & 0.320 & \textbf{0.515} & 0.364 & \textbf{0.175} & 0.101 & \textbf{0.370} \\
\multirow{-21}{*}{\begin{tabular}[c]{@{}c@{}}Math-\\  benchmark\end{tabular}} & \multirow{-7}{*}{\begin{tabular}[c]{@{}c@{}}Qwen2.5-\\  Math-7B\end{tabular}} & \cellcolor[HTML]{DDEBF7}FedGRPO & \cellcolor[HTML]{DDEBF7}0.738 & \cellcolor[HTML]{DDEBF7}0.321 & \cellcolor[HTML]{DDEBF7}0.504 & \cellcolor[HTML]{DDEBF7}\textbf{0.371} & \cellcolor[HTML]{DDEBF7}0.167 & \cellcolor[HTML]{DDEBF7}\textbf{0.110} & \cellcolor[HTML]{DDEBF7}0.369 \\ 

\specialrule{0.8pt}{2pt}{2pt}

 &  & Zero-shot & 0.316 & 0.081 & 0.272 & 0.203 & 0.074 & 0.034 & 0.163 \\
 &  & Fedpetuning+GRPO & 0.348 & 0.114 & 0.272 & 0.210 & 0.049 & 0.040 & 0.172 \\
 &  & Fedpetuning+SFT & 0.352 & 0.092 & 0.262 & 0.223 & 0.060 & 0.031 & 0.170 \\
 &  & DPSDA-FL+GRPO & 0.574 & 0.162 & 0.318 & 0.242 & 0.043 & 0.034 & 0.229 \\
 &  & DPSDA-FL+SFT & 0.480 & 0.110 & 0.281 & 0.259 & 0.055 & 0.042 & 0.205 \\
 &  & Central-GRPO  & 0.724 & 0.257 & 0.392 & 0.338 & \textbf{0.114} & \textbf{0.069} & 0.316 \\
 & \multirow{-7}{*}{\begin{tabular}[c]{@{}c@{}}Qwen2.5-\\  Math-1.5B\end{tabular}} & \cellcolor[HTML]{DDEBF7}FedGRPO & \cellcolor[HTML]{DDEBF7}\textbf{0.740} & \cellcolor[HTML]{DDEBF7}\textbf{0.290} & \cellcolor[HTML]{DDEBF7}\textbf{0.414} & \cellcolor[HTML]{DDEBF7}\textbf{0.353} & \cellcolor[HTML]{DDEBF7}0.099 & \cellcolor[HTML]{DDEBF7}0.066 & \cellcolor[HTML]{DDEBF7}\textbf{0.327} \\
\cmidrule(lr{0.5em}){2-10}

 &  & Zero-shot & 0.118 & 0.092 & 0.063 & 0.046 & 0.006 & 0.006 & 0.055 \\
 &  & Fedpetuning+GRPO & 0.400 & 0.125 & 0.152 & 0.127 & 0.017 & 0.010 & 0.139 \\
 &  & Fedpetuning+SFT & 0.362 & 0.055 & 0.138 & 0.092 & 0.013 & 0.007 & 0.111 \\
 &  & DPSDA-FL+GRPO & 0.392 & 0.154 & 0.127 & 0.117 & \textbf{0.043} & \textbf{0.016} & 0.142 \\
 &  & DPSDA-FL+SFT & 0.416 & 0.121 & 0.171 & 0.169 & 0.016 & 0.013 & 0.151 \\
 &  & Central-GRPO & 0.532 & \textbf{0.248} & \textbf{0.305} & 0.235 & 0.018 & 0.011 & 0.225 \\
 & \multirow{-7}{*}{Qwen2.5-3B} & \cellcolor[HTML]{DDEBF7}FedGRPO & \cellcolor[HTML]{DDEBF7}\textbf{0.546} & \cellcolor[HTML]{DDEBF7}0.235 & \cellcolor[HTML]{DDEBF7}0.296 & \cellcolor[HTML]{DDEBF7}\textbf{0.240} & \cellcolor[HTML]{DDEBF7}0.028 & \cellcolor[HTML]{DDEBF7}0.015 & \cellcolor[HTML]{DDEBF7}\textbf{0.227} \\
\cmidrule(lr{0.5em}){2-10}

 &  & Zero-shot & 0.426 & 0.121 & 0.326 & 0.163 & 0.111 & 0.048 & 0.199 \\
 &  & Fedpetuning+GRPO & 0.504 & 0.129 & 0.317 & 0.181 & 0.121 & 0.039 & 0.215 \\
 &  & Fedpetuning+SFT & 0.497 & 0.143 & 0.297 & 0.189 & 0.072 & 0.008 & 0.201 \\
 &  & DPSDA-FL+GRPO & 0.551 & 0.172 & 0.308 & 0.262 & 0.017 & 0.014 & 0.221 \\
 &  & DPSDA-FL+SFT & 0.574 & 0.239 & 0.351 & 0.256 & 0.087 & 0.048 & 0.259 \\
 &  & Central-GRPO & 0.755 & 0.325 & 0.529 & 0.370 & 0.180 & 0.113 & 0.379 \\
\multirow{-21}{*}{OpenR1-Math} & \multirow{-7}{*}{\begin{tabular}[c]{@{}c@{}}Qwen2.5-\\ Math-7B\end{tabular}} & \cellcolor[HTML]{DDEBF7}FedGRPO & \cellcolor[HTML]{DDEBF7}\textbf{0.768} & \cellcolor[HTML]{DDEBF7}\textbf{0.337} & \cellcolor[HTML]{DDEBF7}\textbf{0.533} & \cellcolor[HTML]{DDEBF7}\textbf{0.382} & \cellcolor[HTML]{DDEBF7}\textbf{0.184} & \cellcolor[HTML]{DDEBF7}\textbf{0.126} & \cellcolor[HTML]{DDEBF7}\textbf{0.388} \\
\bottomrule 
\end{tabular}%
\kern-5pt
\caption{Performance results of FedGRPO and other methods with two trainsets on 1.5B, 3B and 7B models.}
\label{tab:main_results}
\kern-14pt
\end{table*}

This section presents the empirical analysis of the proposed FedGRPO from comparison with baselines, communication efficiency and ablation studies.

\subsection{Experimental Setup}
\label{sec: experimental setup}

\paragraph{Datasets \& Models.} 
We conduct experiments on two classic kinds of LLM tasks including
math problem-solving and question-answering:
\begin{itemize}
    \item For math problem-solving task,  we use MATH-benchmark \cite{hendrycks2021measuring} and OpenR1-Math \cite{openr1} as training sets, and three different sizes of LLMs (Qwen2.5-3B \cite{qwen2.5}, Qwen2.5-Math-1.5B and Qwen2.5-Math-7B \cite{yang2024qwen2}) as the server's foundation models. We evaluate the performance of the model on six widely used math problem test sets, including MATH500 \cite{hendrycks2021measuring}, Minerva \cite{lewkowycz2022solving}, OlympiadBench \cite{he2024olympiadbench}, AIME 2024, AIME 2025, and AMC \cite{li2024numinamath}.
    
    \item For question-answering task, we apply 2WikiMultiHopQA \cite{ho2020constructing} as training set and two LLMs (Qwen2.5-1.5B,Qwen2.5-3B \cite{qwen2.5}) as server's foundation models. The performance is evaluated on the test sets of HotpotQA \cite{yang2018hotpotqa}, 2WikiMultiHopQA \cite{ho2020constructing} and MuSiQue \cite{trivedi2022musique} . Due to space limit, the experimental results of question-answering task are left in Appendix.
\end{itemize}

Each result presented is the average of three repeated experiments. For efficiency, 2,000 samples are randomly selected from the training set for each experiment.


\paragraph{Setting.}
This paper considers two settings for the dataset $\mathcal{D}_s$: 1) none of the $K$ clients possess the ground-truth label $y(x_s)$ for samples in $\mathcal{D}_s$, so clients rely solely on model-based evaluation (see result in Sect. \ref{sec:Performance without Answers}); 2) some clients have access to the ground-truth label $y(x_s)$, and thus can perform answer-based evaluation (see result in Sect. \ref{sec:main_performance}).
We evaluate scenarios involving 4 to 20 clients in federated learning (FL), with up to 320 communication rounds. FedGRPO is assessed under both IID and Non-IID conditions. For the Non-IID setting, we simulate data heterogeneity using the Dirichlet distribution, $\mathrm{Dir}(\beta)$, with $\beta = 0.1$. 

\paragraph{Baselines.}

To benchmark FedGRPO against other client-to-server knowledge transfer approaches in federated foundation models (FedFMs), we consider two representative methods: (1) \textbf{Fedpetuning} \cite{zhang2023fedpetuning} (model-level transfer): clients perform parameter-efficient local tuning and upload only the adapted modules or adapters to the server; (2) \textbf{DPSDA-FL} \cite{abacha2024synthetic} (synthetic data-level transfer): clients use a small foundation model to generate differentially private synthetic data from their local datasets, which are then sent to the server for centralized training. For both baselines, we evaluate models trained with both GRPO and supervised fine-tuning (SFT) objectives.

Furthermore, we compare FedGRPO with the following baselines: (3) \textbf{Zero-shot}: the direct performance of pre-trained LLMs with zero sample post-training, serving as a baseline to measure improvement; (4) \textbf{Central-GRPO} \cite{shao2024deepseekmath}: the group relative policy optimization (GRPO) method proposed by Deepseek, applied in a centralized setting where all training data are aggregated, representing the ideal upper bound for FedGRPO.

\paragraph{Implementation \& Evaluation Metric.}  
The experiments are configured with a learning rate of $3.0 \times 10^{-6}$ and a temperature of $0.7$. Each question involved sampling {8 candidate policies} for evaluation, with the maximum generation length limited to {2048 tokens}. The auxiliary data owned by server has 100 samples and expert selection parameter $L$ is set as 20, $M$ is 2. More settings can be seen in Appendix.

We adopt Pass@1 accuracy on the test sets as our primary evaluation metric, following the methodology of~\cite{shao2024deepseekmath} and a higher value indicates better performance. Additionally, we assess the communication overhead of the complete training process, measured in megabytes (MB), as an efficiency metric and the lower value means more efficiency.








\subsection{Comparison with Baselines}
\label{sec:main_performance}

We conduct a comprehensive performance evaluation of our proposed \textbf{FedGRPO} framework against a range of baselines, including FedFMs methods (model-level transfer method {FedPETuning} \cite{zhang2023fedpetuning} and synthetic  data-level transfer method {DPSDA-FL} \cite{abacha2024synthetic}) and the centralized GRPO method. The detailed results, presented in Table~\ref{tab:main_results}, are evaluated across three model scales (1.5B, 3B, and 7B). Our analysis confirms the superior performance and effectiveness of {FedGRPO}.

\paragraph{Comparison with FedFMs Baselines.}
When benchmarked against existing FedFMs methods, {FedGRPO} consistently demonstrates significant and stable performance gains. Our approach substantially outperforms both {Fedpetuning} (model-level transfer) and {DPSDA-FL} (synthetic data-level transfer), irrespective of their underlying optimization objectives (GRPO or SFT). This superiority holds across all model sizes and both datasets. For instance, using the Qwen2.5-Math-7B model on the Math-benchmark, {FedGRPO} attains an average accuracy of 0.369, marking a substantial improvement over the best-performing federated baseline, DPSDA-FL+SFT (0.253). The performance advantage is also evident on smaller models; for the Qwen2.5-Math-1.5B model on Math-benchmark, {FedGRPO} (0.338) is clearly ahead of the next best federated method, DPSDA-FL+GRPO (0.275). These robust results validate that {FedGRPO} provides a more effective framework for leveraging clients’ domain knowledge to enhance the performance of FMs  compared to other FedFMs methods.

\paragraph{Comparison with Centralized GRPO.}
We compare our method against Centralized GRPO denoted as "{Central-GRPO}" in Tab.\ref{tab:main_results}, which represents a upper bound method trained on aggregated data. The results compellingly demonstrate that {FedGRPO} not only approaches but in several instances surpasses this centralized baseline. For example, on the OpenR1-Math dataset with the 7B model, {FedGRPO} achieves an average accuracy of 0.388, exceeding {Central-GRPO}'s score of 0.379. A similar trend is observed with the 1.5B model on the same dataset. On the Math-benchmark, {FedGRPO} (0.369) performs almost identically to {Central-GRPO} (0.370) with the 7B model. 
This near-centralized performance demonstrates that our competence-based expert selection and dual evaluation mechanisms synergistically produce reliable reward signals, enabling effective and privacy-preserving server model optimization.

\begin{table*}[!ht]
\centering
\small
\renewcommand{\arraystretch}{1.0}
\begin{tabular}{@{}clccccccc@{}}
\toprule
\rowcolor[HTML]{E2EFDA} 
\multicolumn{1}{!{\vrule width 0pt}c}{\cellcolor[HTML]{D0CECE}Model} & 
\multicolumn{1}{c}{\cellcolor[HTML]{D0CECE}Method} & 
Math500                          & 
Minerva                       & 
AMC                           & 
Olympiad                      & 
AIME24                      & 
AIME25                      & 
\multicolumn{1}{c!{\vrule width 0pt}}{\cellcolor[HTML]{FFF2CC}Avg.}  \\ \midrule
                                                                                & Zero-shot                                          & 0.316                         & 0.081                         & 0.272                         & 0.203                         & 0.074                         & 0.034                         & 0.163                         \\
                                                                                & \cellcolor[HTML]{FFFFFF}Central-GRPO              & 0.701 & 0.307 & 0.440 & 0.353 & 0.081 & 0.056 & 0.323                      \\
\multirow{-3}{*}{\begin{tabular}[c]{@{}c@{}}Qwen2.5-Math-1.5B\end{tabular}} & \cellcolor[HTML]{DDEBF7}FedGRPO                    & \cellcolor[HTML]{DDEBF7}0.681 & \cellcolor[HTML]{DDEBF7}0.272 & \cellcolor[HTML]{DDEBF7}0.414 & \cellcolor[HTML]{DDEBF7}0.317 & \cellcolor[HTML]{DDEBF7}0.114 & \cellcolor[HTML]{DDEBF7}0.061 & \multicolumn{1}{c!{\vrule width 0pt}}{\cellcolor[HTML]{DDEBF7}0.310} \\ \midrule
                                                                                & Zero-shot                                          & 0.118                         & 0.092                         & 0.063                         & 0.046                         & 0.006                        & 0.006                         & 0.055                        \\
                                                                                & \cellcolor[HTML]{FFFFFF}Central-GRPO               & 0.474                         & 0.210                        & 0.157                        & 0.153                        & 0.010                        & 0.003                        & 0.168                        \\
\multirow{-3}{*}{Qwen2.5-3B}                                                    & \cellcolor[HTML]{DDEBF7}FedGRPO                    & \cellcolor[HTML]{DDEBF7}0.360  & \cellcolor[HTML]{DDEBF7}0.272 & \cellcolor[HTML]{DDEBF7}0.142 & \cellcolor[HTML]{DDEBF7}0.098 & \cellcolor[HTML]{DDEBF7}0.005 & \cellcolor[HTML]{DDEBF7}0.003 & \multicolumn{1}{c!{\vrule width 0pt}}{\cellcolor[HTML]{DDEBF7}0.146} \\ \midrule
                                                                                & Zero-shot                                          & 0.426                         & 0.121                         & 0.326                         & 0.163                        & 0.111                        & 0.048                         & 0.199                         \\
                                                                                & \cellcolor[HTML]{FFFFFF}Central-GRPO               & 0.742                         & 0.320                        & 0.515                        & 0.364                        & 0.175                         & 0.101                         & 0.370                        \\
\multirow{-3}{*}{\begin{tabular}[c]{@{}c@{}}Qwen2.5-Math-7B\end{tabular}}   & \cellcolor[HTML]{DDEBF7}FedGRPO                    & \cellcolor[HTML]{DDEBF7}0.692 & \cellcolor[HTML]{DDEBF7}0.282 & \cellcolor[HTML]{DDEBF7}0.417 & \cellcolor[HTML]{DDEBF7}0.328 & \cellcolor[HTML]{DDEBF7}0.182 & \cellcolor[HTML]{DDEBF7}0.061 & \multicolumn{1}{c!{\vrule width 0pt}}{\cellcolor[HTML]{DDEBF7}0.327} \\ \bottomrule
\end{tabular}%
\kern-6pt
\caption{Performance results of FedGRPO without ground-truth answers on Math-benchmark dataset.}
\label{tab:without_answer_result}
\kern-14pt
\end{table*}

\subsection{Performance Without Answers}
\label{sec:Performance without Answers}
We assess FedGRPO's robustness and effectiveness when ground-truth answers are unavailable. In this setting, clients employ locally trained reward models to evaluate policies and assign rewards, while FedGRPO selects appropriate rewards via competence-based expert selection. Specifically, we use a locally trained Qwen2.5-Math-1.5B model as the reward model under Non-IID $\beta=0.1$; further details are provided in the Appendix. Notably, other baselines such as Fedpetuning and DPSDA-FL are inapplicable in this scenario due to the absence of reference answers.

As the results shown in Tab.\ref{tab:without_answer_result}, {FedGRPO} demonstrates strong performance even in the absence of ground-truth answers. For the 1.5B model, FedGRPO achieves an average accuracy of {0.310}, which is only slightly lower than the centralized upper bound (Central-GRPO, {0.323}), and significantly outperforms the zero-shot baseline ({0.163}). Similar trends are observed for larger models: with Qwen2.5-Math-7B, FedGRPO attains an average accuracy of {0.327}, closely matching Central-GRPO's {0.369}, and for Qwen2.5-3B, FedGRPO achieves {0.146} compared to Central-GRPO's {0.168}.

\subsection{Communication Efficiency} 
\begin{figure}[!htbp]
    \flushleft 
    \includegraphics[scale=0.6]{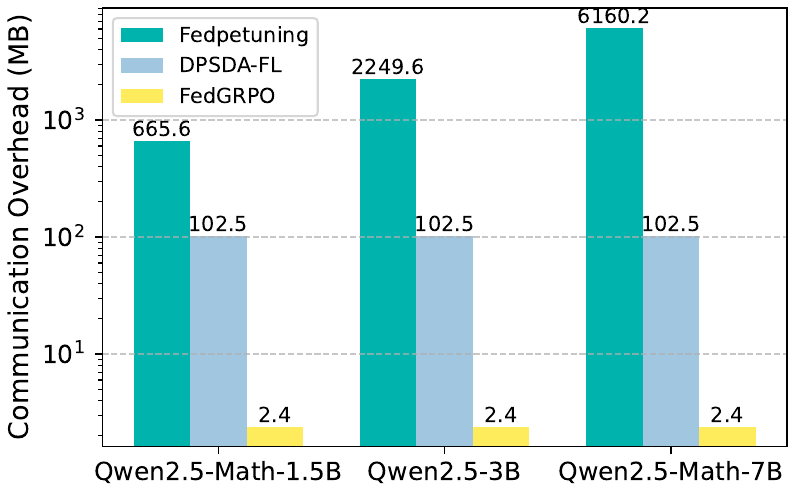}
    \kern-6pt
    \caption{The communication overheaf of FedGRPO, Fedpetuning and DPSDA-FL. }
    \label{fig:overhead}
    \kern-4pt
\end{figure}

To evaluate FedGRPO's communication efficiency, we tested the communication overhead of FedGRPO (transmitting group rewards), Fedpetuning (transmitting LoRA trainable parameters), and DPSDA-FL (transmitting synthetic data) during a complete training process on 1.5B, 3B, and 7B models. The results are shown in Fig.\ref{fig:overhead}.

From the results we can observe that FedGRPO requires only \textbf{2.4~MB} of reward signal transfer during FL, and remains constant regardless of model size because it only transmits a short reward value (1.0 or 0.0) for each policy. This is in stark contrast to the baselines: {DPSDA-FL} requires 102.5~MB (a 40$\times$ increase), while {FedPETuning}'s overhead is even more prohibitive, scaling with the model size to a massive 6.1~GB for the 7B model. This represents a reduction in communication cost of two to three orders of magnitude, making FedGRPO a highly practical and scalable solution for real-world deployment.

\subsection{Ablation Study}
\label{sec: effectiveness sensitive}

\subsubsection{Client Number.}
\begin{figure}[htbp]
    \kern-2pt
    \centering
    \begin{subfigure}[b]{0.48\linewidth}
        \centering
        \includegraphics[width=\textwidth]{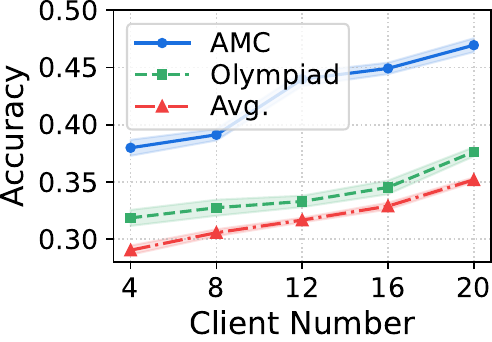} 
        \caption{Qwen2.5-Math-1.5B}
        
        \kern-4pt
    \end{subfigure}%
    \hfill
    \begin{subfigure}[b]{0.48\linewidth}
        \centering
        \includegraphics[width=\textwidth]{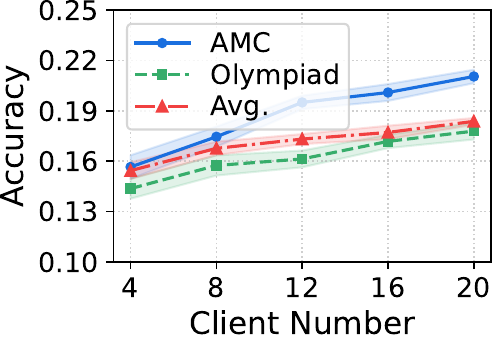} 
        \caption{Qwen2.5-3B}

        \kern-4pt
    \end{subfigure}
    \caption{Accuracy of FedGRPO on AMC, Olympiad, and all 6 testsets (averaged) across varying client numbers.}
    \label{fig:abalation_client_num}
    \kern-12pt
\end{figure}

We evaluate the performance of FedGRPO with client number $K$ ranging from 4 to 20.
As shown in Fig.~\ref{fig:abalation_client_num}, FedGRPO demonstrates consistent performance improvements as more clients participate. For the Qwen2.5-Math-1.5B model (Fig.~\ref{fig:abalation_client_num}a), the average accuracy increases steadily from approximately 0.29 with 4 clients to 0.36 with 20 clients. Notably, the accuracy on the AMC subset rises from about 0.37 to 0.47, and Olympiad accuracy improves from 0.32 to 0.38 as the client number grows. A similar trend is observed for the Qwen2.5-3B model (Fig.~\ref{fig:abalation_client_num}b).  These consistent gains across different benchmarks confirm FedGRPO’s ability to robustly aggregate distributed knowledge and benefit from larger federated networks.

\subsubsection{Expert Selection.}

We evaluated the impact of different selected experts number $M$ on the performance of FedGRPO when performing competence-based expert selection. Due to space limit, the detail results are shown in Appendix.

\section{Conclusion}
\label{sec: conclusion}
In this work, we proposed FedGRPO, a privacy-preserving and communication-efficient framework for Federated Foundation Models (FedFMs), which leverages client-side domain expertise through reinforcement learning. By reformulating large model adaptation as a reward-based evaluation process, FedGRPO introduces two key innovations: competence-based expert selection and a federated Group-Relative Policy Optimization mechanism that aggregates scalar feedback instead of high-dimensional parameters or numerous synthetic data. These designs collectively ensure accurate evaluation, preserve user privacy, and support scalable deployment across heterogeneous clients. Empirical results show that FedGRPO not only closely matches centralized GRPO performance and surpasses other FedFMs methods, but also mitigates privacy risks in other FedFMs methods with significantly reduced communication overhead.

\section*{Acknowledgements}
This work is supported by research fund of Tsinghua University - Tencent Joint Laboratory for Internet Innovation Technology.

\bibliography{aaai2026}

\clearpage
\appendix


\section{Implementation Details}
\label{sec: appendix_settings}
\begin{table*}[!h]
\centering
\caption{Performance results for Qwen2.5-1.5B and Qwen2.5-3B models. Acc and F1 scores on three test sets are reported. The best results for each model group are in bold.}
\small
\renewcommand{\arraystretch}{0.95}
\begin{tabular}{clccccccc}
\toprule

\cellcolor[HTML]{D9D9D9} & \multicolumn{1}{c}{\cellcolor[HTML]{D9D9D9}} 
& \multicolumn{2}{c}{\cellcolor[HTML]{E2EFDA}2WikiMultiHopQA} 
& \multicolumn{2}{c}{\cellcolor[HTML]{DDEBF7}HotpotQA} 
& \multicolumn{2}{c}{\cellcolor[HTML]{E6E6FF}MuSiQue} 
& \cellcolor[HTML]{FFF2CC} \\
\multirow{-2}{*}{\cellcolor[HTML]{D9D9D9}Model} 
& \multicolumn{1}{c}{\multirow{-2}{*}{\cellcolor[HTML]{D9D9D9}Method}} 
& \cellcolor[HTML]{E2EFDA}Acc & \cellcolor[HTML]{E2EFDA}F1 
& \cellcolor[HTML]{DDEBF7}Acc & \cellcolor[HTML]{DDEBF7}F1 
& \cellcolor[HTML]{E6E6FF}Acc & \cellcolor[HTML]{E6E6FF}F1 
& \multirow{-2}{*}{\cellcolor[HTML]{FFF2CC}\begin{tabular}[c]{@{}c@{}}Avg \\ Acc\end{tabular}} \\

& Zero-shot & 0.117 & 0.126 & 0.063 & 0.061 & 0.061 & 0.042 & 0.080 \\
& Fedpetuning+GRPO & 0.185 & 0.235 & 0.177 & 0.239 & 0.085 & 0.104 & 0.132 \\
& Fedpetuning+SFT & 0.192 & 0.258 & 0.211 & 0.248 & 0.091 & 0.125 & 0.165 \\
& DPSDA-FL+GRPO & 0.254 & 0.288 & 0.264 & 0.275 & 0.101 & 0.115 & 0.196 \\
& DPSDA-FL+SFT & 0.272 & 0.325 & 0.288 & 0.384 & 0.112 & 0.175 & 0.224 \\
& Central-GRPO & \textbf{0.293} & \textbf{0.503} & 0.313 & 0.462 & 0.125 & 0.188 & 0.239 \\
\multirow{-7}{*}{Qwen2.5-3B} & \cellcolor[HTML]{DDEBF7}FedGRPO & \cellcolor[HTML]{DDEBF7}0.285 & \cellcolor[HTML]{DDEBF7}0.498 & \cellcolor[HTML]{DDEBF7}\textbf{0.322} & \cellcolor[HTML]{DDEBF7}\textbf{0.480} & \cellcolor[HTML]{DDEBF7}\textbf{0.128} & \cellcolor[HTML]{DDEBF7}\textbf{0.192} & \cellcolor[HTML]{DDEBF7}\textbf{0.245} \\
\cmidrule(lr{0.5em}){2-9} 

& Zero-shot & 0.013 & 0.012 & 0.017 & 0.010 & 0.005 & 0.004 & 0.070 \\
& Fedpetuning+GRPO & 0.104 & 0.135 & 0.098 & 0.121 & 0.015 & 0.019 & 0.072 \\
& Fedpetuning+SFT & 0.128 & 0.144 & 0.114 & 0.147 & 0.021 & 0.032 & 0.084 \\
& DPSDA-FL+GRPO & 0.205 & 0.125 & 0.211 & 0.234 & 0.033 & 0.045 & 0.150 \\
& DPSDA-FL+SFT & 0.224 & 0.158 & 0.238 & 0.249 & 0.041 & 0.049 & 0.163 \\
& Central-GRPO & 0.271 & \textbf{0.241} & \textbf{0.277} & 0.254 & 0.069 & 0.072 & 0.206 \\
\multirow{-7}{*}{Qwen2.5-1.5B} & \cellcolor[HTML]{DDEBF7}FedGRPO & \cellcolor[HTML]{DDEBF7}\textbf{0.276} & \cellcolor[HTML]{DDEBF7}0.237 & \cellcolor[HTML]{DDEBF7}0.272 & \cellcolor[HTML]{DDEBF7}\textbf{0.261} & \cellcolor[HTML]{DDEBF7}\textbf{0.071} & \cellcolor[HTML]{DDEBF7}\textbf{0.073} & \cellcolor[HTML]{DDEBF7}\textbf{0.206} \\

\bottomrule 
\end{tabular}%
\label{tab:qa_results_final}
\end{table*}

\paragraph{FedGRPO Hyperparameters.}
For the FedGRPO experiments, we set the communication epoch to 320 and use the Adam optimizer with a learning rate of $3 \times 10^{-6}$. The coefficient for the group-relative reward is set to 8, while the coefficient for the format reward is set to 1. All experiments are conducted on 8 NVIDIA A100 GPUs using BF16 (Brain Floating Point) precision. The prompt template used for training is as follows: 

\begin{tcolorbox}[
    center,
    arc=0mm,
    boxrule=1pt,
    colback=blue!6!white,
    colframe=black,
    colbacktitle=black,
]

\textbf{System Prompt:} You are a helpful AI Assistant that provides well-reasoned and detailed responses. For a given question, you should first think the reasoning process in the mind step by step and then provide the user with the answer. The reasoning process and answer must be enclosed within \verb|<think>| \verb|</think>| and \verb|<answer>|\verb|</answer>|, respectively.You must use the following format:“\verb|<think>\n| {put your think process here} \verb|</think>\n| \verb|<answer>\n| {put your answer here} \verb|</answer>\n|”

\textbf{User:} This is the problem: \verb|{QUESTION}| \\
\textbf{Assistant:} \verb|...|
\end{tcolorbox}

\paragraph{Performance without Answers (Section 4.3).}
In the experiments described in Section 4.3 (\emph{Performance without Answers}), each client trains a Qwen2.5-Math-1.5B model as its reward model using the GRPO method. The training settings for these experiments are kept consistent with those used for FedGRPO in Section 4.1.

\paragraph{Supervised Fine-Tuning (SFT) Training.}
For the Supervised Fine-Tuning (SFT) objective, we set the learning rate to $1.0 \times 10^{-5}$. The prompt template used for SFT training is as follows:

\begin{tcolorbox}[
    center,
    arc=0mm,
    boxrule=1pt,
    colback=blue!6!white,
    colframe=black,
    colbacktitle=black,
]

\textbf{User:} Below is an instruction that describes a task. Write a response that appropriately completes the request.

Instruction:
\verb|{INSTRUCTION}|

Response:
\verb|{RESPONSE}|

\textbf{Assistant:} \verb|...|
\end{tcolorbox}



\section{Experiments on Question-Answer Task}
\label{sec: appendix_QA}

This section makes the comparison between FedGRPO with other methods  in Question-Answer Task. 

\paragraph{Datasets and Models:}
We apply two LLMs including Qwen2.5-1.5B, Qwen2.5-3B \cite{qwen2.5} as server's foundation models. 
For model training, we split the 2WikiMultiHopQA dataset~\cite{ho2020constructing} into training and test sets, and train our models on the training set. For evaluation, we not only use the test split of 2WikiMultiHopQA, but also employ HotpotQA \cite{yang2018hotpotqa} and MuSiQue \cite{trivedi2022musique} as additional test sets. 

\paragraph{Evaluation Metric:}
We use \textbf{accuracy} to measure strict string match with golden answers and \textbf{F1 score} to quantify partial matching through token overlap between predicted and golden answers \cite{li2025r3}.

\paragraph{Experimental Analysis:} We can draw two conlusion from Tab. \ref{tab:qa_results_final}: 1) FedGRPO demonstrates a marked improvement over both zero-shot inference and conventional federated fine-tuning baselines across three challenging QA benchmarks. Specifically, in the Qwen2.5-3B setting, FedGRPO achieves an average accuracy of 24.5\%, representing a more than three-fold gain over the zero-shot baseline (8.0\%) and a substantial margin above federated SFT (16.5\%). Similarly, in the Qwen2.5-1.5B configuration, FedGRPO attains 20.6\% average accuracy versus 7.0\% for zero-shot and 8.4\% for federated SFT. Notably, FedGRPO also outperforms the DPSDA-FL+SFT approach—which relies on synthetic data augmentation—with absolute gains of approximately 2.1\% and 4.3\% in the 3B and 1.5B models, respectively. These results underscore the efficacy of the group-relative reward aggregation mechanism in distilling domain expertise from private clients without direct data sharing.

2) FedGRPO narrows the gap with, and in some cases slightly surpasses, centralized GRPO optimization. Specifically, for Qwen2.5-3B, FedGRPO’s average accuracy of 24.5\% edges out the centralized GRPO result of 23.9\%, while for Qwen2.5-1.5B both methods converge to 20.6\%. On the individual tasks, FedGRPO attains near-state-of-the-art F1 scores on \textsc{2WikiMultiHopQA} (0.498 vs.\ 0.503 for centralized) and exceeds centralized performance on \textsc{HotpotQA} F1 (0.480 vs.\ 0.462). Although absolute scores remain modest on \textsc{MuSiQue}, FedGRPO still leads all federated baselines (F1 = 0.192). Collectively, these findings validate that FedGRPO’s privacy-preserving, reinforcement-learning-inspired protocol can harness heterogeneous client feedback to substantially boost multi-hop and compositional QA performance in large pretrained models.

\section{Ablation Study of Expert Selection}
\label{sec:appendix_ablation}

\begin{figure}[htbp]

    \centering
    \begin{subfigure}[b]{0.48\linewidth}
        \centering
        \includegraphics[width=\textwidth]{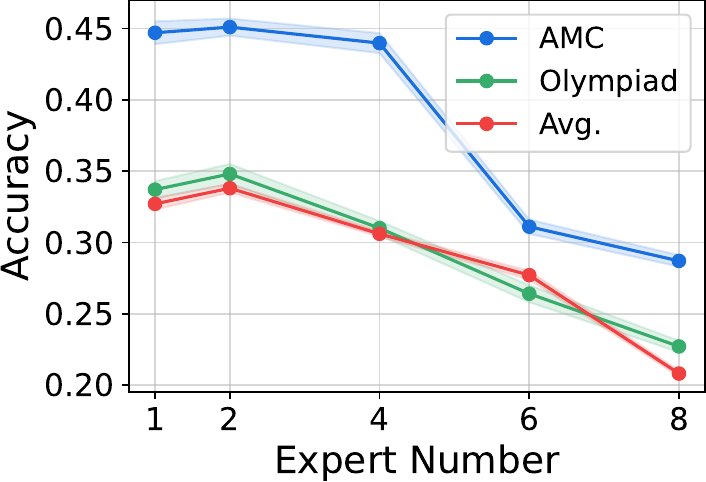} 
        \caption{Qwen2.5-Math-1.5B}
        \label{fig: runtime}
    \end{subfigure}%
    \hfill
    \begin{subfigure}[b]{0.48\linewidth}
        \centering
        \includegraphics[width=\textwidth]{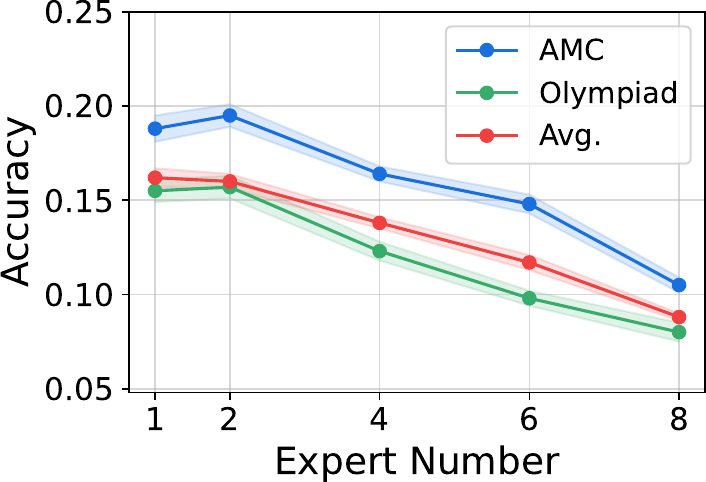} 
        \caption{Qwen2.5-3B}
        \label{fig: flops}
    \end{subfigure}
    \caption{Performance of FedGRPO on AMC, Olympiad and averaged accuracy on 6 testsets including MATH500, Minerva, OlympiadBench, AIME 2024, AIME 2025, and AMC with different selected expert numbers.}
    \label{fig:abalation_expert_num}
\end{figure}

This section investigates the impact of the number of selected experts \(M\) on the overall performance of FedGRPO on AMC, Olympiad and averaged accuracy on 6 testsets. We conduct experiments by varying \(M\) in \(\{1,2,4,6,8\}\) from total 8 clients while keeping all other hyperparameters fixed.  Figure~\ref{fig:abalation_expert_num} indicates that selecting a moderate number of experts (e.g., \(M=2\)) strikes the best balance between capturing heterogeneous client expertise and avoiding noisy or redundant signals. In practice, this choice leads to stable convergence and maximizes the utility of the group‐relative reward aggregation mechanism. Future work may explore adaptive schemes to dynamically adjust \(M\) based on client confidence or task complexity.

\end{document}